\documentclass{article}
\usepackage{spconf,amsmath,graphicx}

\usepackage{enumitem}
\setlist{nosep, leftmargin=14pt}

\usepackage{mwe} 

\usepackage{bm}
\usepackage[table,xcdraw]{xcolor}
\usepackage{amssymb}
\usepackage{adjustbox}
\usepackage{multirow}

\let\OLDthebibliography\thebibliography
\renewcommand\thebibliography[1]{
  \OLDthebibliography{#1}
  \setlength{\parskip}{0pt}
  \setlength{\itemsep}{1pt plus 0.3ex}
}

\newcommand\blfootnote[1]{%
\begingroup
\renewcommand\thefootnote{}\footnote{#1}%
\addtocounter{footnote}{-1}%
\endgroup
}

\title{RECIST Weakly Supervised Lesion Segmentation via Label-Space Co-Training}
\name{Lianyu Zhou\textsuperscript{1}, Dong Wei\textsuperscript{2}, Donghuan Lu\textsuperscript{2}, Wei Xue\textsuperscript{3}, Liansheng Wang\textsuperscript{1}\sthanks{Correspondence: lswang@xmu.edu.cn}, Yefeng Zheng\textsuperscript{2}}
\address{\textsuperscript{1}Xiamen University; \textsuperscript{2}Tencent Jarvis Lab; \textsuperscript{3}OPPO}
\begin{document}
%
\maketitle
\begin{abstract}
As an essential indicator for cancer progression and treatment response, tumor size is often measured following the response evaluation criteria in solid tumors (RECIST) guideline in CT slices.
By marking each lesion with its longest axis and the longest perpendicular one,  laborious pixel-wise manual annotation can be avoided.
However, such a coarse substitute cannot provide a rich and accurate base to allow versatile quantitative analysis of lesions.
To this end, we propose a novel weakly supervised framework to exploit the existing rich RECIST annotations for pixel-wise lesion segmentation.
Specifically, a pair of under- and over-segmenting masks are constructed for each lesion based on its RECIST annotation and served as the label for co-training a pair of subnets, respectively, along with the proposed label-space perturbation induced consistency loss to bridge the gap between the two subnets and enable effective co-training.
Extensive experiments are conducted on a public dataset to demonstrate the superiority of the proposed framework regarding the RECIST-based weakly supervised segmentation task and its universal applicability to various backbone networks.
\blfootnote{L. Zhou, D. Wei and D. Lu contributed equally.}
\end{abstract}
\section{Introduction}
\label{sec:intro}

Tumor size measurement in medical imaging and follow-ups is a widely accepted protocol for cancer monitoring
\cite{beaumont2019radiology}.
In current clinical routine, the measurement is often performed in computed tomography (CT) slices manually by trained specialists, following the response evaluation criteria in solid tumors (RECIST) guideline \cite{EISENHAUER2009NEW}.
Specifically, a couple of RECIST diameters are marked for each lesion (Fig. \ref{fig: framework}(b)), with the major diameter measuring the longest axis of the lesion and the minor measuring the longest perpendicular axis to the major, in the axial slice where the lesion appears largest.
The RECIST diameters are commonly adopted as a time-efficient alternative to full lesion segmentation (e.g., stroking along the lesion boundary precisely or pixel-wise annotation), i.e., a trade-off between measurement accuracy and annotation effort.
As a coarse substitute, however, they cannot provide as much information as the full segmentation, which would allow versatile quantitative analysis such as area estimation or morphological properties.

Despite its great value, manual segmentation of medical images is notoriously laborious, tedious, error-prone, and subject to inter-observer variability.
Therefore, tremendous efforts have been made to develop automated methods \cite{wei2015medical}.
Significant breakthroughs have been achieved in the past decade by deep convolutional neural networks (DCNNs) \cite{hesamian2019deep}.
However, full supervision of the DCNNs requires pixel-wise annotations of large datasets \cite{wang2020conquering}, 
which can be difficult or costly to obtain in practice.
To reduce this burden, weak supervision in the form of simplified or partial labels, e.g., bounding boxes \cite{dai2015boxsup} or scribbles \cite{lin2016scribblesup}, 
has been studied with great interests.
For lesion segmentation, the existing RECIST annotations are a natural and rich source of weak supervision.

Recently, several works have emerged that utilized the RECIST annotations for weakly supervised lesion segmentation \cite{RN21,cai2018accurate,li2020deep,tang2020one,tang2021lesion,tang2021weakly}.
Despite the remarkable progress, we identify three common drawbacks of existing works that should be overcome to push the research towards better clinical applicability.
First, almost all the existing methods \cite{RN21,cai2018accurate,li2020deep,tang2020one}
relied on the GrabCut algorithm \cite{rother2004grabcut}---a classical unsupervised segmentation algorithm---to generate initial pseudo ground truth from the RECIST diameters for training the segmentation networks.
As the quality of the pseudo ground truth largely depended on the initial seeds for GrabCut, the seeding strategy must be carefully
devised to optimize the pseudo ground truth, which may encounter difficulty in generalization in practice; besides, the delicate pre-generation stage added unnecessary complexity to the training process, and even so, these empirical strategies were not guaranteed to always generate lesion masks with high fidelity.
Second, many existing works involved iterative training procedures, 
where the pseudo ground truth was updated and the network performance gradually improved in rounds \cite{cai2018accurate,li2020deep}, leading to exceedingly time-consuming training process.
%
Third, most previous works assumed that the lesion-of-interest (LOI; an enlarged bounding box of the lesion) region was cropped out beforehand as input \cite{RN21,cai2018accurate,li2020deep,tang2021weakly}.
Such assumption, while acceptable when developing early-stage prototypes, may pose an obstacle to practical application as the clinicians still need to obtain the LOI first.
A straightforward solution is to prefix a lesion detection model to the segmentation network \cite{tang2020one}.
However, such a two-stage structure can be unnecessarily redundant if a simpler one-stage model can achieve the same performance.

In this work, we present a novel one-stage co-training framework for RECIST supervised lesion segmentation in CT slices, which effectively addresses all the three drawbacks.
Rather than over-tuning a classical unsupervised method for optimal pseudo ground truth, we instead make an intuitive observation that the RECIST diameters naturally compose two sets of masks: a quadrilateral connecting, and a circle circumscribing the four end points.
Thus, we train a model whose two subnets are supervised with either of the two masks, respectively,
Noting that the two masks are inherently under- and over-segmented representations of the lesion, respectively, we obtain the final prediction by averaging the corresponding under- and over-segmenting predictions.
In this way, the training label construction is simple and straightforward.
Inspired by the recent progress in self supervision with contrastive learning \cite{he2020momentum}, 
we propose a novel consistency loss that contrasts the two subnets' predictions for label-space perturbation based co-training \cite{blum1998combining}.
Owing to the simplicity, robustness, and efficacy of the proposed dual label construction and label-space co-training, our framework can accept whole CT slices as input
and get rid of the iterative refinement, while still being able to produce high-quality lesion segmentation.
Last but not least, our framework is model-agnostic and can be readily applied to any standard segmentation backbone.
Experimental results on a public dataset demonstrate the advantages of our framework over existing approaches.\vspace{-2mm}

\section{Method}
\label{sec:format}


\noindent{\bf Problem Formulation.}
Following the literature, the target of this work is to train a model with RECIST annotations to perform accurate dense pixel-wise classification of lesion and non-lesion on axial CT slices.
Formally, we view a slice and its lesion mask as two $K$-dimensional vectors:
$\bm{I} \in [0,1]^K$ and $\bm{M} \in \{0,1\}^K$, respectively, where all pixels of a slice constitute an index set $\mathcal{P}=\{p|p=0,1,\dots,K-1\}$.
For a pixel $p$, $\bm{M}_p=1$ indicates that it is foreground (a lesion pixel), whereas $\bm{M}_p=0$ indicates background;
all foreground pixels constitute a set $\mathcal{M}=\{p|\bm{M_p}=1,p\in \mathcal{P}\}$.
Hence, $\bm{M}$ and $\mathcal{M}$ are alternative representations of the same mask.
Similarly, the RECIST annotation of a lesion can be represented as a mask $\bm{R}$ and a corresponding index set $\mathcal{R}$,
where $\bm{R}_p=1$ indicates the pixel $p$ is on the RECIST diameters and 0 otherwise.
Therefore, given a training slice set with RECIST annotations\footnote{A slice may contain multiple lesions and thus has a set of annotations.} $D^\mathrm{train}=\{(\bm{I}, \{\bm{R}\})\}$, the target is to train a segmentation model that can predict accurate $\bm{\hat{M}}$ for any unannotated test slice.

\begin{figure}[!t]
\centering
\includegraphics[width=\linewidth,trim=0 2 0 1,clip]{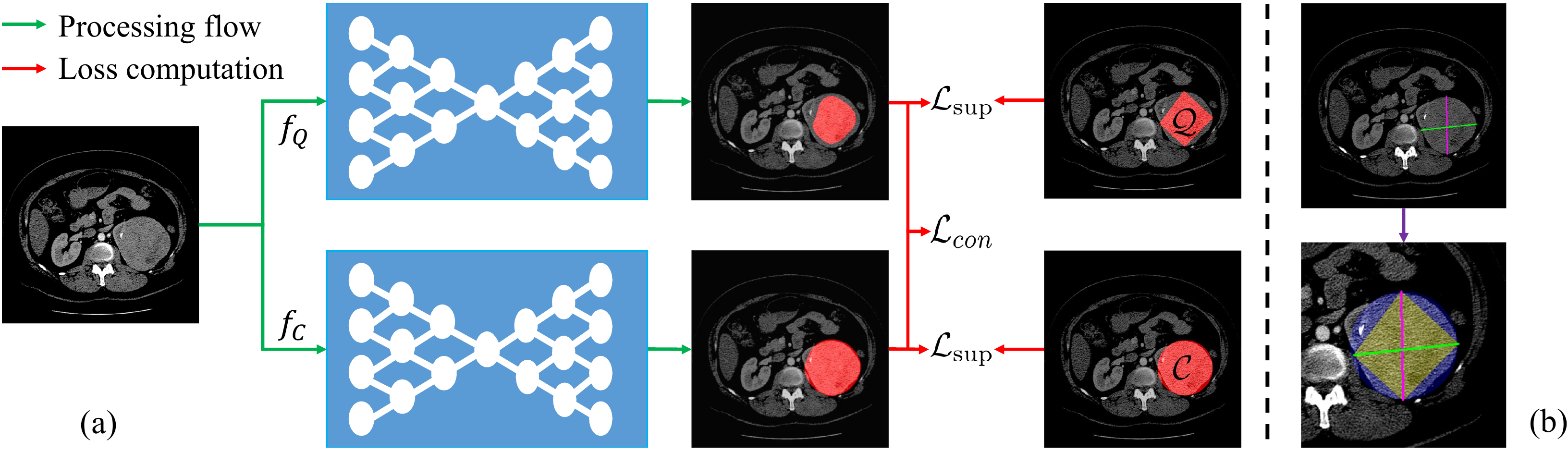}\vspace{-2mm}
\caption{(a) Overview of the proposed framework.
(b) Illustration of the dual mask construction.
Top: a CT slice with RECIST diameters overlaid.
Bottom: the same slice zoomed for better visibility, where yellow color indicates definite foreground (the quadrilateral $\mathcal{Q}$), blue indicates the uncertain area $\mathcal{A}$, their combination indicates the circumscribed circle $\mathcal{C}$, and the rest color-less region indicates definite background.}
\label{fig: framework}\vspace{-2.5mm}
\end{figure}

\noindent\textbf{Dual Mask Construction from RECIST Diameters.}
In this work, we propose straightforward construction of a pair of simple yet contrasting masks for network supervision (Fig. \ref{fig: framework}(a)):
1) a quadrilateral that connects the four endpoints of the RECIST diameters with enclosing pixels forming an index set $\mathcal{Q}$ for mask $\bm{Q}$: $\mathcal{Q}=\{p|\bm{Q}_p=1, p\in\mathcal{P}\}$, and
2) the minimum circumscribed circle 
of the diameters with enclosing pixels forming an index set $\mathcal{C}$ for mask $\bm{C}$: $\mathcal{C}=\{p|\bm{C}_p=1, p\in\mathcal{P}\}$.
Since most RECIST-measurable lesions show convex shapes \cite{EISENHAUER2009NEW}, it is reasonable to analyze the general properties of $\bm{Q}$ and $\bm{C}$ assuming lesions with convex outlines \cite{zlocha2019improving}.
Based on this assumption, it is easy to see that the  quadrilateral is completely contained in the lesion, i.e., $\mathcal{Q}\subseteq\mathcal{M}$.
Therefore, $\bm{Q}$ is an under-segmentation of the genuine lesion mask $\bm{M}$.
On the contrary, $\bm{C}$ is an over-segmentation of $\bm{M}$, i.e., $\mathcal{M}\subseteq\mathcal{C}$.
In fact, this circle is also the minimum circumscribed circle of the lesion, i.e., the smallest circle that can fully contain the lesion while including as little background as possible, given the lesion is convex.
This is favorable especially compared to the bounding boxes of the RECIST diameters \cite{yan2019mulan}, which guarantees neither full inclusion of the lesion nor least inclusion of background.
With the obvious ``flaws'' of being under- and over-segmenting, 
directly training with either of these two seemingly naive masks would lead to suboptimal results.
Next, we describe how to effectively utilize them to build our framework.

\noindent\textbf{Co-Training with Label-Space Perturbation Induced Consistency Loss.}
As show in Fig. \ref{fig: framework}(a), our framework mainly consists of two subnetworks $f_Q$ and  $f_C$ supervised by $\bm{Q}$ and $\bm{C}$, respectively, with its loss function defined as:
\begin{equation}
    \mathcal{L}_\mathrm{sup} = \ell_{\mathcal{P}}\big(\bm{\hat{Q}},\bm{Q}\big) +
    \ell_{\mathcal{P}}\big(\bm{\hat{C}}, \bm{C}\big),
    \label{eq: Spv}
\end{equation}
where $\bm{\hat{Q}}=f_Q(\bm{I})$ and $\bm{\hat{C}}=f_C(\bm{I})$ are the predicted probability maps by the two networks for a slice $\bm{I}$, respectively, and $\ell_\mathcal{P}(\cdot, \cdot)$ is a segmentation loss (such as the Dice loss) effective in
set $\mathcal{P}$.
$\mathcal{L}_\mathrm{sup}$ drives $f_Q$ and $f_C$ to output predictions mimicking their supervision masks $\bm{Q}$ and $\bm{C}$, respectively.
Therefore, the predictions are expected to be under- and over-segmenting.
A natural approach to fusing them is an averaging ensemble:
$\bm{\hat{M}} = {(\bm{\hat{Q}}+\bm{\hat{C}})}/{2}$.
As later shown in our experiment, $\bm{\hat{M}}$ is consistently better than both $\bm{\hat{Q}}$ and $\bm{\hat{C}}$.
This is reasonable: as $\bm{\hat{Q}}$ and $\bm{\hat{C}}$ are biased estimations of the underlying ground truth towards diverging directions, averaging them is supposed to effectively cancel the biases.
However, such a setting is essentially a simple ensemble of two networks trained synchronously but independently.
To bridge the two networks, we propose to co-train $f_Q$ and $f_C$ by explicitly enforcing the consistency between $\bm{\hat{Q}}$ and $\bm{\hat{C}}$ with a contrasting loss:\vspace{-3.5mm}
\begin{equation}
    \mathcal{L}_\mathrm{con}=\ell_{\mathcal{P}}\big(\bm{\hat{Q}}, \bm{\hat{C}} \big).
    \label{eq: Con}
\end{equation}
The rationale is that $\bm{Q}$ and $\bm{C}$ are in fact intentionally introduced perturbations to the same underlying ground truth masks.
Enforcing the consistency in the two subnets' predictions across the perturbations can make them gradually approach a proper balance between the under- and over-segmenting biases, thus producing a more accurate segmentation.
With a weight factor $\lambda$, combining $\mathcal{L}_\mathrm{sup}$ with $\mathcal{L}_\mathrm{con}$ yields the complete optimization target of our framework:
\begin{equation}
        \mathcal{L}=\ell_{\mathcal{P}}\big(\bm{\hat{Q}}, \bm{Q} \big) +
        \ell_{\mathcal{P}}\big(\bm{\hat{C}}, \bm{C} \big) +
        \lambda \ell_{\mathcal{P}}\big(\bm{\hat{Q}}, \bm{\hat{C}} \big).
\label{eq: basic loss}
\end{equation}

\noindent\textbf{Region-Constrained Consistency Loss.}
Referring to Fig. \ref{fig: framework}(b), it is easy to derive that:
$\forall p\in    \mathcal{Q},\, \bm{Q}_p=\bm{M}_p=\bm{C}_p=1$,
and $\forall p\notin \mathcal{C},\, \bm{Q}_p=\bm{M}_p=\bm{C}_p=0$,
thus we have
\begin{equation}
        \forall p \in \mathcal{Q}\cup \big(\mathcal{P}-\mathcal{C}\big),\,
        \bm{Q}_p=\bm{M}_p=\bm{C}_p.
\end{equation}
This is to say, both $\bm{Q}$ and $\bm{C}$ are consistent with the underlying ground truth lesion mask $\bm{M}$ in regions inside $\mathcal{Q}$ (definite foreground) and outside $\mathcal{C}$ (definite background).
In our design, $\mathcal{L}_\mathrm{con}$ is targeted at the ambiguous region where $\bm{Q}$ and $\bm{C}$ disagree and the real ground truth is unknown, which is the region outside $\mathcal{Q}$ but inside $\mathcal{C}$: $\mathcal{A}=\mathcal{C}-\mathcal{Q}$.
On one hand, when $f_Q$ and $f_C$ are trained well by $\bm{Q}$ and $\bm{C}$, respectively, their predictions should be the same in the agreed region $\mathcal{Q}\cup(\mathcal{P}-\mathcal{C})$, and $\mathcal{L}_\mathrm{con}$ does not function.
On the other hand, if either one of them makes a mistake in this region, the mistake may mislead the other network through $\mathcal{L}_\mathrm{con}$.
Therefore, we further propose to improve $\mathcal{L}_\mathrm{con}$ by constraining its effective region within $\mathcal{A}$ instead of the whole slice $\mathcal{P}$,
thus Eq. (\ref{eq: basic loss}) becomes:
\begin{equation}
    \mathcal{L}=\ell_{\mathcal{P}}\big(\bm{\hat{Q}}, \bm{Q} \big) +
    \ell_{\mathcal{P}}\big(\bm{\hat{C}}, \bm{C} \big) +
    \lambda \ell_{\mathcal{A}}\big(\bm{\hat{Q}}, \bm{\hat{C}} \big),
\label{eq: advanced loss}
\end{equation}
which is expected to be more efficient, effective, and robust.

\noindent\textbf{Relation with Related Work.}
Conventional co-training often co-trained the subnets with distinct inputs (e.g., different views of a web page \cite{blum1998combining} or MRI sequences in an exam \cite{wang2021acn}).
In this work, in an attempt to utilize the inherent uncertainty of the RECIST annotation, we instead co-train the two subnets with two distinct supervision masks which are purposely constructed with diverging biases.
This is in contrast with the GrabCut-based methods which tried to increase the certainty of the supervising mask via engineering with empirical rules.
To this end, our work presents a new perspective of utilizing the RECIST annotation for weak supervision.
It is worth noting that our label-space perturbation is also inspired by the data-space perturbations (i.e., data augmentation) commonly used in self-supervised contrastive learning \cite{he2020momentum}.

\section{Experiments and Results}
\label{sec:pagestyle}

\noindent\textbf{Dataset.} 
We evaluate the proposed framework on the publicly available
2019 Kidney Tumor Segmentation (KiTS19) challenge data \cite{heller2019kits19},
which provides kidney tumor masks of 210 abdominal CT scans of unique patients in arterial phase.
We randomly split the 210 volumes into training and test sets in the ratio of 80:20.
For minimal preprocessing, the slices are rescaled to the range [0, 1] with a soft-tissue CT window range of $[0, 400]$ Hounsfield unit (HU), and resized to $512\times512$ pixels.
For weak supervision,
we follow \cite{cai2018accurate,li2020deep} to convert the annotation masks to RECIST diameters by measuring the major and minor axes on 2D slices.
The Dice score, Jaccard index, and 95\textsuperscript{th} percentile of the Hausdorff distance (HD95) are used as evaluation metrics as in \cite{luo2021semi}.

\begin{table}[t]\vspace{-2.5mm}
\caption{Training configurations.
}\label{tab:train_cfg}
\centering
\begin{adjustbox}{width=.975\linewidth}
\begin{tabular}{ccccc}
\hline\rowcolor[HTML]{EFEFEF}
Backbone       & U-Net              & HNN~\cite{cai2018accurate} & ARU-Net~\cite{tang2020one} & Swin Transformer~\cite{LIU2021SWIN}  \\ \hline
Pretrain       & None               & None                       & None                       & ADE20K~\cite{zhou2017scene}          \\
Batch size     & 6                  & 16                         & 10                         & 6                                    \\
Prepare epochs & 250                & 250                        & 300                        & 150                                  \\
Total epochs   & 600                & 600                        & 1000                       & 600                                  \\
Optimizer      & \multicolumn{3}{c}{AdaMax~\cite{kingma2014adam}}                             & AdamW~\cite{loshchilov2017decoupled} \\
Learning rate  & 1$\times$10$^{-3}$ & 1$\times$10$^{-3}$         & 1$\times$10$^{-3}$         & 6$\times$10$^{-5}$                   \\
Scheduler      & None               & None                       & None                       & None                                 \\
GPU            & \multicolumn{4}{c}{NVIDIA RTX 2080TI$\times$1}                                                                      \\
Augmentation   & \multicolumn{4}{c}{Flipping, cropping, padding, rotating, and scaling}                                                 \\ \hline
\end{tabular}
\end{adjustbox}\vspace{-2.5mm}
\end{table}

\noindent\textbf{Implementation.}
All experiments are conducted with PyTorch 1.7.1 \cite{paszke2019pytorch}.
For backbone, we mainly consider
the U-Net \cite{RONNEBERGER2015U} and Swin Transformer \cite{LIU2021SWIN} (base).
The former is arguably the mostly widely used architecture for medical image segmentation, and the latter is a recent SOTA model for general image segmentation.
We also implement our framework with two other backbones advocated in related works: the holistically nested networks (HNNs) \cite{cai2018accurate} and ARU-Net \cite{tang2020one}, to demonstrate its model-agnostic applicability.
As our purpose is to validate the effectiveness of the proposed dual-label consistency loss, no advanced post-processing trick is implemented;
instead, we simply apply a default threshold of 0.5
to identify lesion pixels.
The soft Dice loss \cite{MILLETARI2016V} is used for $\ell_\mathcal{P}$ and $\ell_\mathcal{A}$.
We first train the model with only the pseudo-label supervision by $\mathcal{L}_\mathrm{sup}$ until convergence (the ``preparation''), 
and then add the consistency loss $\mathcal{L}_\mathrm{con}$ for remaining epochs.
This is because $\bm{\hat{Q}}$ and $\bm{\hat{C}}$ should be reasonable under- and over-segmentations, respectively, for $\mathcal{L}_\mathrm{con}$ to contrast them validly.
More details about the training configurations
are charted in Table \ref{tab:train_cfg}.
Our codes will be released.

\begin{table}[t]\vspace{-2.5mm}
\caption{Lesion segmentation performance 
    and comparison to SOTA approaches (mean$\pm$margin of error at $95\%$ confidence level).
    The strong baseline \cite{zlocha2019improving} and upper bound are fully supervised models trained with GrabCut-generated and ground truth masks, respectively.
    A fixed $\lambda=0.4$ is used for our framework, given its robustness to different $\lambda$ values (cf. Fig. \ref{fig:lambda}).
    *:  significance at 0.05 level for pairwise comparison to our framework with the same backbone.
    }
    \label{tab:results_n_compare}
  \centering
  \setlength{\tabcolsep}{.5mm}
  \begin{adjustbox}{width=\linewidth}
  \begin{tabular}{ccccc}
\hline\rowcolor[HTML]{EFEFEF}
Method                      & Backbone           & Dice $\uparrow$          & Jaccard $\uparrow$       & HD95 (pixel) $\downarrow$ \\ \hline
Baseline                    & U-Net              & 0.854$\pm$0.025          & 0.778$\pm$0.028          & 4.083$\pm$0.204           \\
WSSS \cite{cai2018accurate} & HNN                & 0.838$\pm$0.030          & 0.766$\pm$0.032          & 4.118$\pm$0.233*          \\
DRL \cite{li2020deep}       & U-Net+Deep Q-net   & 0.858$\pm$0.023          & 0.779$\pm$0.025          & 3.963$\pm$0.181           \\
RLS \cite{tang2021weakly}   & U-Net              & 0.838$\pm$0.021*         & 0.744$\pm$0.023*         & 3.989$\pm$0.185           \\
MULAN \cite{yan2019mulan}   & Mask R-CNN         & 0.878$\pm$0.022          & 0.808$\pm$0.025          & 3.718$\pm$0.199           \\
SEENet \cite{tang2020one}   &\small Mask R-CNN+ARU-Net & 0.890$\pm$0.020          & 0.825$\pm$0.023          & 3.516$\pm$0.191           \\
\multirow{3}{*}{Ours}       & HNN                & 0.845$\pm$0.028          & 0.770$\pm$0.029          & 3.889$\pm$0.180           \\
                            & U-Net              & 0.862$\pm$0.026          & 0.792$\pm$0.028          & 3.955$\pm$0.206           \\
                            & ARU-Net            & 0.894$\pm$0.019          & 0.827$\pm$0.022          & 3.482$\pm$0.161           \\
Upper bound                 & U-Net              & 0.866$\pm$0.029          & 0.805$\pm$0.030          & 3.758$\pm$0.225*          \\ \hline
\rowcolor[HTML]{EFEFEF}
Method                      & Backbone           & Dice $\uparrow$          & Jaccard $\uparrow$       & HD95 (pixel) $\downarrow$ \\ \hline
Baseline                    & Swin-T             & 0.891$\pm$0.018*         & 0.822$\pm$0.021*         & 3.449$\pm$0.144           \\
RLS \cite{tang2021weakly}   & Swin-T             & 0.888$\pm$0.012*         & 0.808$\pm$0.015*         & 3.354$\pm$0.116           \\
Ours                        & Swin-T             & \textbf{0.907}$\pm$0.016 & \textbf{0.846}$\pm$0.020 & \textbf{3.316}$\pm$0.158  \\
Upper bound                 & Swin-T             & 0.912$\pm$0.018          & 0.856$\pm$0.021          & 3.312$\pm$0.117           \\ \hline
\end{tabular}
\end{adjustbox}\vspace{-2mm}
\end{table}

\begin{figure}[t]
  \centering
  \includegraphics[width=.71\linewidth,trim=0 3 0 0,clip]{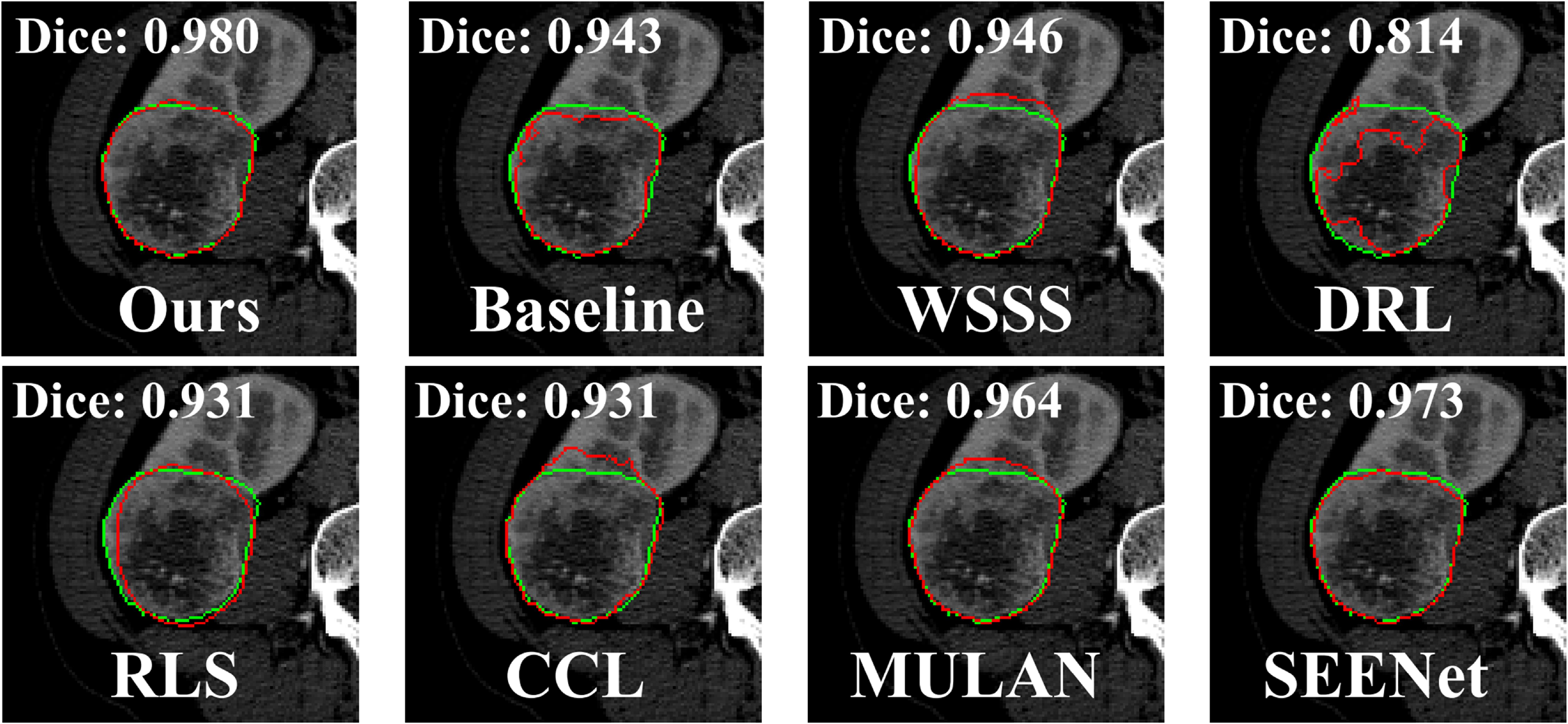}\vspace{-2mm}
  \caption{Segmentation results of different methods (green: ground truth; red: network prediction; zoomed for better visibility).
  Baseline, RLS, and our framework use the Swin Transformer backbone, whereas the others
  use the backbone networks suggested in the original papers.}
  \label{fig: vis-compare}\vspace{-2.5mm}
\end{figure}

\noindent\textbf{Dual Mask Validation.}
First, we verify the fundamental assumption that the constructed dual masks $\bm{Q}$
and $\bm{C}$
can serve as under- and over-segmenting supervision, respectively,
by evaluating their recall and precision against the ground truth $\bm{M}$. 
For ideal under segmentation, the recall should be low while the precision should be high, and vice versa for over segmentation.
The recall and precision are 0.715 and 0.990 for $\bm{Q}$, and 0.982 and 0.802 for $\bm{C}$,
verifying the assumption.
Besides, compared to fitting an ellipse to the RECIST diameters~\cite{tang2021weakly} (recall 0.954 and precision 0.854), our $\bm{C}$ is a better over segmentation as indicated by its apparently higher recall and lower precision,
thus more suitable for our framework.

\noindent\textbf{Lesion Segmentation.}
We evaluate the segmentation performance of our proposed framework and compare to several SOTA approaches.
The evaluation is done with whole-slice input (instead of cropped LOIs) for more practical use scenarios, albeit more challenging.
For compared methods, we mainly use the backbones suggested in the original papers and follow the optimal training schedules described thereby;
otherwise the U-Net and Swin Transformer are adopted and optimized for model-agnostic approaches.
The results
are shown in Table \ref{tab:results_n_compare}.
As expected, the choice of the backbone network has a major impact on the performance, with the Swin Transformer performing the best.
Despite that, when the same backbones are used, our framework not only improves upon the strong baseline in all metrics, but also generally outperforms all competing methods.
In fact, the Dice scores of our framework are fairly close to those of the upper bound (trained with pixel-wise full supervision by ground truth masks) using the same backbones.
These results demonstrate the effectiveness and robustness of our proposed framework, although it is simpler in terms of both mask initialization (compared to the GrabCut) and network design (compared to the DRL \cite{li2020deep} and SEENet \cite{tang2020one}).
Fig. \ref{fig: vis-compare} shows example segmentation results by different methods.

\begin{table}[t]\vspace{-2.5mm}
\caption{Ablation study (with U-Net) results in Dice scores (mean$\pm$ margin of error at $95\%$ confident level).}
\label{tab:ablate}
\centering
\begin{adjustbox}{width=.7\linewidth}
\begin{tabular}{cc|ccc}
\hline
\rowcolor[HTML]{EFEFEF}
$\ell_{\mathcal{P}}$ & $\ell_{\mathcal{A}}$ & $\bm{\hat{Q}}$  & $\bm{\hat{C}}$  & ${(\bm{\hat{Q}} + \bm{\hat{C}})}/{2}$ \\ \hline
$\times$             & $\times$             & 0.683$\pm$0.023 & 0.809$\pm$0.024 & 0.826$\pm$0.023                               \\
\checkmark           & $\times$             & 0.781$\pm$0.024 & 0.829$\pm$0.022 & 0.848$\pm$0.022                               \\
$\times$             & \checkmark           & 0.844$\pm$0.022 & 0.852$\pm$0.021 & \textbf{0.862}$\pm$0.026                      \\ \hline
\end{tabular}
\end{adjustbox}\vspace{-2mm}
\end{table}

\begin{figure}[t]
    \centering
    \includegraphics[width=.725\linewidth,height=2.6cm,trim=0 15 0 0,clip]{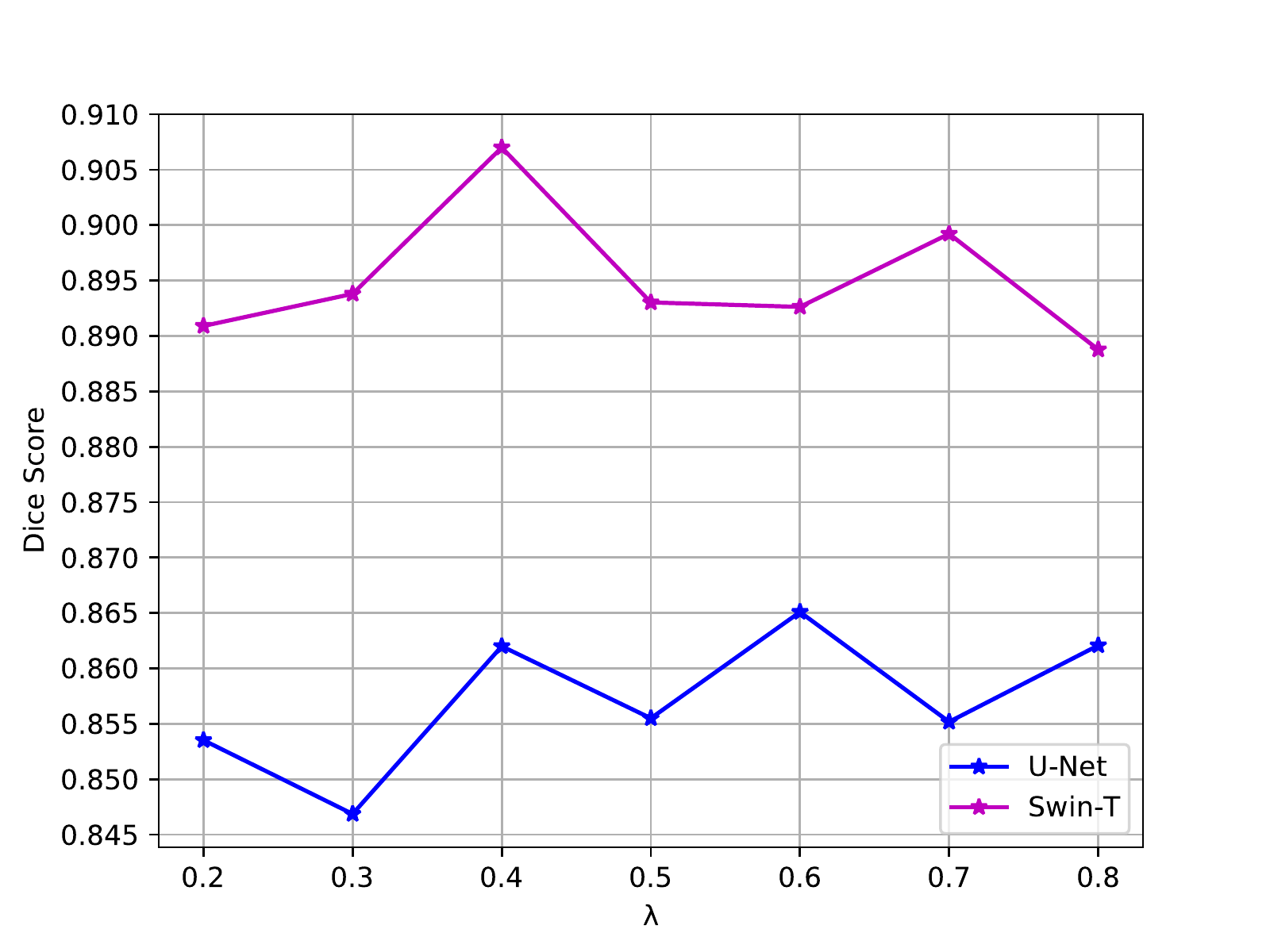}\vspace{-2mm}
    \caption{Effects of varying $\lambda$ ($x$ axis) on performance ($y$ axis). 
    }\label{fig:lambda}\vspace{-2.5mm}
    \end{figure}

\noindent\textbf{Ablation Study.}
We conduct ablation studies with and without co-training by $\mathcal{L}_\mathrm{con}$, and with $\mathcal{L}_\mathrm{con}$ effective on the whole slice ($\ell_{\mathcal{P}}$) and in the constrained region ($\ell_{\mathcal{A}}$).
As shown in Table \ref{tab:ablate},
with $\ell_{\mathcal{P}}$ added, substantial improvements are observed in both $\bm{\hat{Q}}$ and $\bm{\hat{C}}$, as well as the ensemble;
when additionally constraining the effective region of the consistency loss with $\ell_{\mathcal{A}}$, further improvements are achieved.
Eventually, the ensemble of $\bm{\hat{Q}}$ and $\bm{\hat{C}}$ achieves the best performance with $\ell_{\mathcal{A}}$.
These results demonstrate the efficacy of the proposed label-space co-training framework, especially with the region-constrained consistency loss.
In addition, our framework introduces only one hyperparameter, i.e., $\lambda$ in Eqn. (\ref{eq: advanced loss}) that controls the relative importance of the co-training loss.
As shown in Fig. \ref{fig:lambda}, the small variations (less than 0.025 Dice score) corresponding to both backbones demonstrate that
our framework is not sensitive to the exact value of $\lambda$ in a reasonable range ([0.4, 0.7]).

\section{Conclusion}
This work presented a novel co-training framework for lesion segmentation in CT slices, with weak supervision by the RECIST annotations,
which effectively co-trained two subnets with a novel label-space perturbation induced consistency loss. 
Extensive experiments validated the efficacy of co-training with the proposed consistency loss, our framework's superiority to existing works, and its model-agnostic property.

\vspace{2.5mm}
\noindent\textbf{Compliance with Ethical Standards.}
This research study was conducted retrospectively using human subject data made available in open access by the KiTS19 challenge \cite{heller2019kits19}.
Ethical approval was not required as confirmed by the license attached with the open access data.

\noindent\textbf{Conflicts of Interest.}
The authors have no relevant financial or non-financial interests to disclose.

\noindent\textbf{Acknowledgments.}
This work was supported by the Ministry of Science and Technology of the People's Republic of China (STI2030-Major Projects2021ZD0201900).

%
%
%

\bibliographystyle{IEEEbib}
\bibliography{strings,refs,mybibliography}

\end{document}